\documentclass[final]{cvpr}

\usepackage{times}
\usepackage{epsfig}
\usepackage{graphicx}
\usepackage{amsmath}
\usepackage{amssymb}
\usepackage{enumitem}

\usepackage{booktabs}
\usepackage{multirow}
\usepackage{makecell}

\usepackage{color, colortbl}
\definecolor{iGray}{gray}{0.9}
\definecolor{beaublue}{rgb}{0.74, 0.83, 0.9}
\definecolor{Royal_Blue}{rgb}{0.0, 0.1, 0.66}
\usepackage[pagebackref=true,breaklinks=true,letterpaper=true,colorlinks,bookmarks=false, citecolor = Royal_Blue]{hyperref}
\usepackage{amssymb}
\usepackage{pifont}
\newcommand{\cmark}{\ding{51}}%
\newcommand{\xmark}{\ding{55}}%

\DeclareMathOperator*{\argmax}{arg\,max}
\DeclareMathOperator*{\argmin}{arg\,min}
\usepackage{bbding}

\usepackage{graphicx}
\usepackage{subcaption}
\usepackage{mwe}
\usepackage{amsmath, bm}
\usepackage{float}

\usepackage{booktabs}
\usepackage{multirow}
\usepackage{makecell}

\usepackage[pagebackref=true,breaklinks=true,colorlinks,bookmarks=false]{hyperref}

\def\cvprPaperID{1629} 
\def\confYear{CVPR 2021}

\begin{document}

\title{LightTrack: Finding Lightweight Neural Networks for Object Tracking\\via One-Shot Architecture Search}

\author{Bin Yan$^{1,2,*}$, \href{https://houwenpeng.com/}{\textcolor{black}{Houwen Peng}}$^{1, *, \dagger}$,~Kan Wu$^{1,3,*}$, Dong Wang$^{2, \dagger}$, Jianlong Fu$^{1}$, and Huchuan Lu$^{2,4}$\\
$^1$Microsoft Research Asia~
$^2$Dalian University of Technology\\
$^3$Sun Yat-sen University~
$^4$Peng Cheng Laboratory
}

\maketitle
\pagestyle{empty}
\thispagestyle{empty}


\begin{abstract}
	\vspace{-2mm}

    Object tracking has achieved significant progress over the past few years.  However,  state-of-the-art trackers become increasingly heavy and expensive, which limits their deployments in resource-constrained applications. In this work, we present LightTrack, which uses neural architecture search (NAS) to design more lightweight and efficient object trackers. 
    Comprehensive experiments show that our LightTrack is effective.  It can find trackers that achieve superior performance compared to handcrafted SOTA trackers, such as SiamRPN++~\cite{SiamRPNplusplus} and Ocean~\cite{Ocean}, while using much fewer model Flops and parameters.   Moreover,  when deployed on resource-constrained mobile chipsets, the discovered trackers run much faster. For example, on Snapdragon 845 Adreno GPU, LightTrack runs $12\times$ faster than Ocean, while using $13\times$ fewer parameters and $38\times$ fewer Flops. Such improvements might narrow the gap between academic models and industrial deployments in object tracking task. LightTrack is released at \href{https://github.com/researchmm/LightTrack}{here}.
   
\end{abstract}

\newcommand\blfootnote[1]{%
\begingroup 
\renewcommand\thefootnote{}\footnote{#1}%
\addtocounter{footnote}{-1}%
\endgroup 
}
{
	\blfootnote{
	 $^*$Equal contributions. Work performed when Bin and Kan are interns of MSRA. ~$^\dagger$ Corresponding authors.  
	}
}
\vspace{-4.5mm}
\section{Introduction}
\vspace{-0.5mm}

\begin{figure}[!t]
	\begin{center}
		\includegraphics[width=1\linewidth]{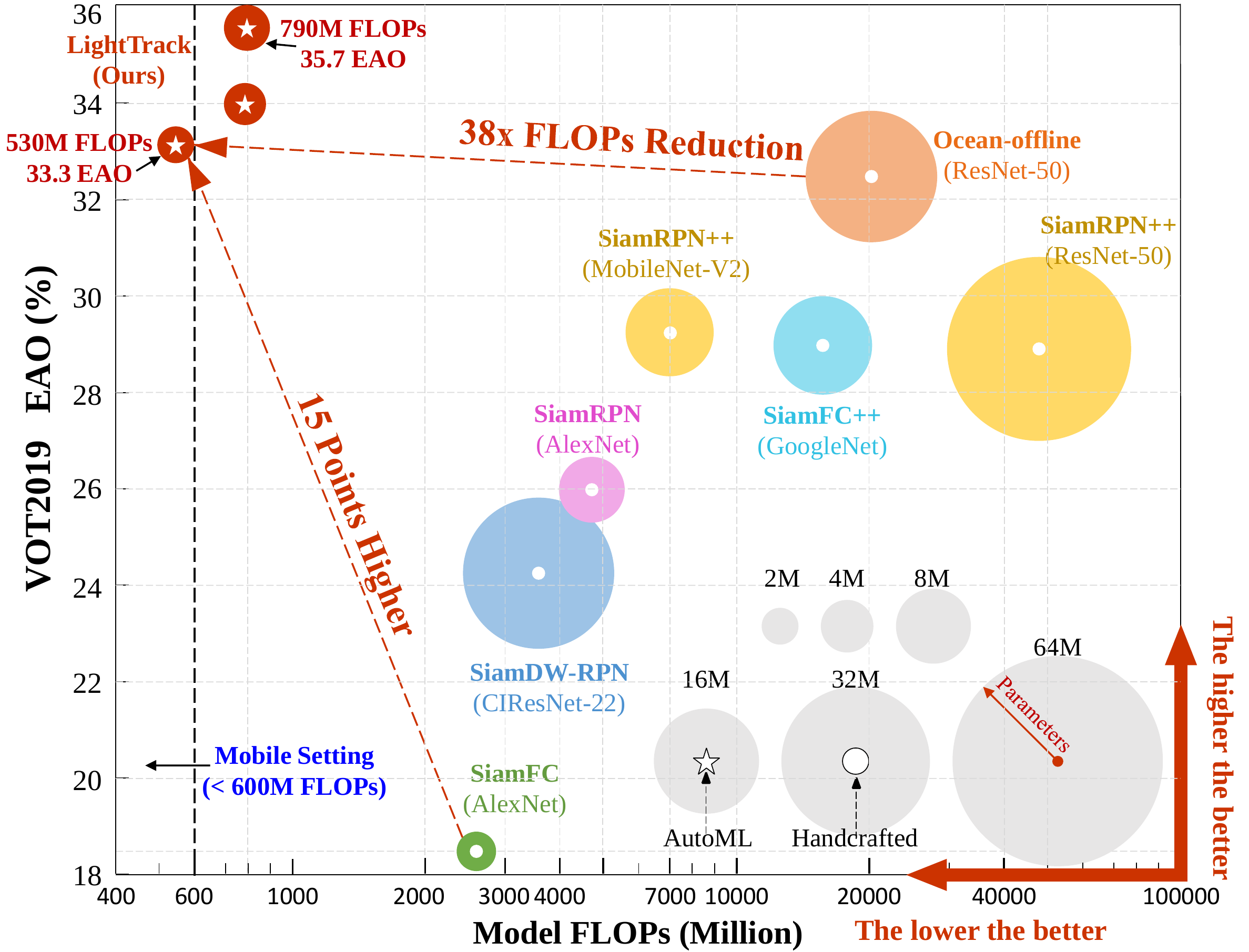}
	\end{center}
	\vspace{-6mm}
	\caption{Comparisons with state-of-the-art trackers in terms of EAO performance, model Flops and parameters on VOT-19 benchmark. The circle diameter is in proportion to the size of model parameter. The proposed LightTrack is superior than SiamFC~\cite{SiameseFC}, SiamRPN~\cite{SiameseRPN}, SiamRPN++~\cite{SiamRPNplusplus}, SiamFC++~\cite{SiamFC++} and Ocean~\cite{Ocean}, while using much fewer Flops and parameters. Best viewed in color.
	}
	\label{fig1}
	\vspace{-5mm}
\end{figure}

Object tracking is one of the most fundamental yet challenging tasks in computer vision. Over the past few years, due to the rise of deep neural networks, object tracking has achieved remarkable progress. But meanwhile, tracking models are becoming increasingly heavy and expensive. For instance, the latest SiamRPN++~\cite{SiamRPNplusplus} and Ocean~\cite{Ocean} trackers respectively utilize 7.1G and 20.3G model Flops as well as 11.2M and 25.9M parameters to achieve state-of-the-art performance, being much more complex than the early SiamFC~\cite{SiameseFC} method (using 2.7G Flops and 2.3M parameters), as visualized in Fig.~\ref{fig1}. Such large model sizes and expensive computation costs hinder the deployment of tracking models in real-world applications, such as camera drones, industrial robotics, and driving assistant system, where model size and efficiency are highly constrained.

There are two straightforward ways to tackle the complexity and efficiency issues. One is model compression, while the other is compact model designing. Existing off-the-shelf compression techniques such as pruning and quantization can reduce model complexity, while they inevitably bring non-negligible performance degradation due to information loss~\cite{deepcompress,TStracker}. 
On the other hand, handcrafting new compact and efficient models is engineering expensive and heavily relies on human expertise and experience~\cite{Deeper-wider-SiamRPN,Survey}.  

This paper introduces a new solution -- automating the design of lightweight models with \emph{neural architecture search} (NAS), such that the searched trackers can be carried out in an efficient fashion on resource-limited hardware platforms. It is non-trivial because that object trackers typically need ImageNet pre-training, while NAS algorithms require the performance feedback on the target tracking task as supervision signals. Based upon recent one-shot NAS~\cite{ENAS,understand_os,SPOS}, we propose a new search algorithm dedicated to object tracking task, called \emph{LightTrack}. It encodes all possible architectures into a backbone supernet and a head supernet. The backbone supernet is pre-trained on ImageNet then fine-tuned with tracking data, while the head supernet is directly trained on tracking data. The supernets are trained only once, then each candidate architecture inherits its weights from the supernets directly. Architecture search is performed on the trained supernets, using tracking accuracy and model complexity as the supervision guidance. On the other hand, to reduce model complexity, we design a search space consisting of lightweight building blocks, such as depthwise separable convolutions~\cite{DSConv} and inverted residual structure~\cite{Mobilenetv2,MobileNetv3}. Such search space allows 
the one-shot NAS algorithm to search for more compact neural architectures, striking a balance between tracking performance and computational costs. 

Comprehensive experiments verify that LightTrack is effective. It is able to search out efficient and lightweight object trackers. For instance, LightTrack finds a 530M Flops tracker, which achieves an EAO of 0.33 on VOT-19 benchmark, surpassing the SOTA SiamRPN++~\cite{SiamRPNplusplus} by 4.6\% while reducing its model complexity (48.9G Flops) by 98.9\%. More importantly, when deployed on resource-limited chipsets, such as edge GPU and DSP, the discovered tracker performs very competitive and runs much faster than existing methods. On Snapdragon 845 Adreno 630 GPU~\cite{snapdragon}, our LightTrack runs $12\times$ faster than Ocean~\cite{Ocean} (38.4 \emph{v.s.} 3.2 \emph{fps}), while using $13\times$ fewer parameters (1.97 \emph{v.s.} 25.9 M) and $38\times$ fewer Flops (530 \emph{v.s.} 20,300 M). Such improvements enable deep tracking models to be easily deployed and run at real-time speed on resource-constrained hardware platforms.

This work makes the following contributions.
\vspace{-2mm}
\begin{itemize}[leftmargin=0.468cm]
	\item{
		We present the first effort on automating the design of neural architectures for object tracking. We develop a new formulation of one-shot NAS and use it to find promising architectures for tracking.
	}
\vspace{-2mm}
	\item{
		We design a lightweight search space and a dedicated search pipeline for object tracking. Experiments verify the proposed method is effective. Besides, the searched trackers achieve state-of-the-art performance and can be deployed on diverse resource-limited platforms.
	}
\end{itemize}	


\section{Related Work}\label{Sec2}

{\textbf{Object Tracking}.} 
In recent years, siamese trackers have become popular in object tracking. The pioneering works are SiamFC and SINT~\cite{SiameseFC,SINT}, which propose to combine naive feature correspondence with the siamese framework. A large number of follow-up works have been proposed and achieved significant improvements~\cite{SiamBAN,GCT,Zuo,TADT,SiamRCNN}. They mainly fall into three camps: more precise box estimation, more powerful backbone, and online update. 
More concretely, in contrast to the multiple-scale estimation in SiamFC, later works like SiamRPN~\cite{SiameseRPN} and SiamFC++~\cite{SiamFC++} leverage either anchor-based or anchor-free mechanism for bounding box estimation, which largely improve the localization precision. Meanwhile, SiamRPN++~\cite{SiamRPNplusplus} and Ocean~\cite{Ocean} take the powerful ResNet-50~\cite{ResNet} instead of AlexNet~\cite{AlexNet} as the backbone to enhance feature representation capability. 
On the other hand, ATOM~\cite{ATOM}, DiMP~\cite{DiMP}, and ROAM~\cite{ROAM} combine online update~\cite{MDNet} with the siamese structure and achieve state-of-the-art performance.

Though these methods achieve remarkable improvements, yet they bring much additional computation workload and large memory footprint, 
thus limiting their usage in real-world applications.
For example, deep learning on mobile devices commonly requires model Flops to be less than 600M Flops~\cite{OFA}, \emph{i.e.}, \emph{mobile setting}. However, SiamRPN++~\cite{SiamRPNplusplus} with ResNet-50 backbone has 48.9G Flops, which exceeds the mobile setting by $\sim$80 times. 
Even SiamFC~\cite{SiameseFC}, using the shallow AlexNet, still cannot satisfy the restricted computation workload when deployed on embedded devices.  
In summary, there is a lack of studies on finding a good trade-off between model accuracy and complexity in object tracking.

\textbf{Neural Architecture Search}. NAS aims at automating the design of neural network architectures. Early methods search a network using either reinforcement learning~\cite{NASRL} or evolution algorithms~\cite{geneticCNN}. These approaches require training thousands of architecture candidates from scratch, leading to unaffordable computation overhead. Most recent works resort to the one-shot weight sharing strategy to amortize the searching cost~\cite{randomNAS,ENAS}. The key idea is to train a single over-parameterized hypernetwork model, and then share the weights across subnets. 
Single-path with uniform sampling~\cite{SPOS} is one representative method in one-shot regime. In each iteration, it only samples one random path and trains the path using one batch data. Once the training process is finished, the subnets can be ranked by the shared weights. On the other hand, instead of searching over a discrete set of architecture candidates, differentiable methods~\cite{DARTS,PDARTS} relax the search space to be continuous, such that the search can be optimized by the efficient gradient descent. Recent surveys on NAS can be found in~\cite{Survey}.

NAS is primarily proposed for image classification and recently extended to other vision tasks, such as image segmentation~\cite{Auto-Deeplab} and object detection~\cite{NAS-FPN}. Our work is inspired by the recent DetNAS~\cite{DetNAS}, but has three fundamental differences. First, the studied task is different. DetNAS is designed for object detection, while our work is for object tracking. 
Second, DetNAS only searches for backbone networks by fixing the head network with a pre-defined handcrafted structure. This may lead to that the searched backbone is sub-optimal, because it is biased towards fitting the fixed head, rather than the target task. In contrast, our method searches backbone and head architectures simultaneously, aiming to find the most promising combination for the target tracking task. Last, the search space is different. We design a new search space for object tracking dedicated to search for lightweight architectures.

\section{Preliminaries on One-Shot NAS}\label{Sec3}
Before introducing the proposed method, we briefly review the one-shot NAS approach, which serves as the basic search algorithm discussed in this work.
One-shot NAS treats all candidate architectures as different subnets of a supernet 
and shares weights between architectures that have common components. More concretely, 
the architecture search space $\mathcal{A}$ is encoded in a supernet, denoted as $\mathcal{N}(\mathcal{A}, W)$, where $W$ is the weight of the supernet. The weight $W$ is shared across all the architecture candidates, \emph{i.e.}, subnets ${\alpha\in \mathcal{A}}$ in $\mathcal{N}$. The search of the optimal architecture $\alpha^{*}$ 
is formulated as a nested optimization problem:
\begin{equation}
	\label{eqn1}
	\begin{aligned}
		\alpha^*  = \mathop{\argmax}_{\alpha\in \mathcal{A}} \emph{Acc}_\emph{val} \left( \mathcal{N}(\alpha, W^*(\alpha)) \right), \\
		\mathrm{s.t.}\quad 
		W^* = \mathop{ \argmin_{W} } \mathcal{L}_{\emph{train}}(\mathcal{N}(\mathcal{A}, W)),
	\end{aligned}
\end{equation}
where the constraint function is to optimize the weight $W$ of the supernet $\mathcal{N}$ by minimizing the loss function  $\mathcal{L}_\emph{train}$ on \emph{training} dataset, while the objective function is to search architectures via ranking the accuracy $\emph{Acc}_\emph{val}$ of subnets on \emph{validation} dataset based on the learned supernet weight $W^*$. 
Only the weights of the single supernet $\mathcal{N}$ need to be trained, and subnets 
can then be evaluated without any separate training by inheriting trained weights from the one-shot supernet. 
This greatly speeds up performance estimation of architectures, since no subnet training is required, resulting in the method only costs a few GPU days.

To reduce memory footprint, one-shot methods usually sample subnets from the supernet $\mathcal{N}$ for optimization. For simplicity, this work adopts the single-path uniform sampling strategy, \emph{i.e.}, each batch only sampling one random path from the supernet for training~\cite{randomNAS,SPOS}. This single-path one-shot method decouples the supernet training and architecture optimization. 
Since it is impossible to enumerate all the architectures ${\alpha\in \mathcal{A}}$ for performance evaluation, we resort to evolutionary  algorithms~\cite{real2019regularized,SPOS} to find the most promising subnet from the one-shot supernet.

\section{LightTrack}\label{Sec4}

Searching lightweight architectures for object tacking is a non-trivial task. There exist three key challenges. 

\vspace{-0.1cm}
\begin{itemize}[leftmargin=0.3cm]
	\item{ First, in general, object trackers need model pre-training on image classification task for a good initialization, while NAS algorithms require supervision signals from target tasks. 
		Searching architectures for object tracking requires to consider both the pre-training on ImageNet and the fine-tuning on tracking data. 
	}
	\vspace{-0.2cm}
	\item{ Second, object trackers usually contain two parts: a backbone network for feature extraction and a head network for object localization. 
	When searching for new architectures, NAS algorithms needs to consider the two parts as a whole, such that the discovered structures are suitable for the target tracking task. 
	}

	\vspace{-0.2cm}
	\item{ Last but not the least, search space is critical for NAS algorithms and it defines which neural architectures a NAS approach might discover in	principle. 
		To find lightweight architectures, the search space requires to include compact and low-latency building blocks.
	}
	\vspace{-0.1cm}
\end{itemize}

In this section, we tackle the aforementioned challenges and propose LightTrack based on one-shot NAS. 
We first introduce a new formulation of one-shot NAS specialized for object tracking task. 
Then, we design a lightweight search space consisting of depthwise separable convolutions~\cite{DSConv} and inverted residual structure~\cite{Mobilenetv2,MobileNetv3}, which allows the construction of efficient tracking architectures. 
At last, we present the pipeline of LightTrack, which is able to search diverse models for different deployment scenarios. 

\subsection{Tracking via One-Shot NAS}
Current prevailing object trackers (such as~\cite{SiameseRPN,ATOM,DiMP}) 
all require ImageNet pre-training for their backbone networks, such that the trackers can obtain good image representation. However, for architecture search, it is impossible to pre-train all backbone candidates individually on ImageNet, because the computation cost is very huge (ImageNet pre-training usually takes several days on 8 V100 GPUs just for a single network). 
Inspired by one-shot NAS, we introduce the weight-sharing strategy to eschew pre-training each candidate from scratch. More specifically, we encode the search space of backbone architectures into a supernet $\mathcal{N}_b$. This backbone supernet only needs to be pre-trained once on ImageNet, and its weights are then shared across different backbone architectures which are subnets of the one-shot model. 
The ImageNet pre-training is performed 
%
by optimizing the classification loss function $\mathcal{L}_{\emph{pre-train}}^{\emph{cls}}$ as
\begin{equation}
	\label{eqn_spos_class}
	W_b^p =  ~\mathop{ \argmin_{{W}_b} } \mathcal{L}_{\emph{pre-train}}^{\emph{cls}}(\mathcal{N}_b(\mathcal{A}_b, W_b)),
\end{equation}
where $\mathcal{A}_b$ represents the search space for backbone architectures, while $W_b$ denotes the parameter of the backbone supernet $\mathcal{N}_b$. The pre-trained weight $W_b^p$ are shared across different backbone architectures and serve as the initialization for the subsequent search of tracking architectures. Such weight-sharing scheme allows the ImageNet pre-training to be performed only on the backbone supernet instead of each subnet, thereby reducing the training costs by orders of magnitude. 

Deep neural networks for object tracking generally contain two parts: one pre-trained backbone network for feature extraction and one head network for object localization. These two parts work together to determine the capacity of a tracking architecture. Therefore, for architecture search, it is critical to search the backbone and head networks as a whole, such that the discovered structure is well-suited to tracking task. To this end, we construct a tracking supernet $\mathcal{N}$ consisting of the backbone part $\mathcal{N}_b$ and the head part $\mathcal{N}_h$, which is formulated as $\mathcal{N} = \{\mathcal{N}_b, \mathcal{N}_h\}$. The backbone supernet $\mathcal{N}_b$ is first pre-trained on ImageNet by Eq.~(\ref{eqn_spos_class}) and generates the weight $W_b^p$. 
The head supernet $\mathcal{N}_h$ subsumes all possible localization networks in the space $\mathcal{A}_h$ and shares the weight $W_b$ across architectures. The joint search of backbone and head architectures is conducted on \emph{tracking} data, which reformulates the one-shot NAS as
\begin{equation}
	\label{eqn_spos_track}
	\begin{aligned}
		\alpha_b^*, \alpha_h^*  = \mathop{\argmax}_{\alpha_b, \alpha_h \in \mathcal{A}}  \emph{Acc}_{\emph{val}}^{\emph{trk}} \left( \mathcal{N}(\alpha_b, W_b^*(\alpha_b); \alpha_h, W_h^*(\alpha_h)) \right), \\
		\mathrm{s.t.}\quad 
		W_b^*, W_h^* = ~\mathop{ \argmin_{W_b \leftarrow W_b^p, W_h} } \mathcal{L}_{\emph{train}}^{\emph{trk}}(\mathcal{N}(\mathcal{A}_b, W_b;\mathcal{A}_h, W_h)), 
	\end{aligned}
\end{equation}
where the constraint function is to train the tracking supernet $\mathcal{N}$ and optimize the weights $W_b$ and $W_h$ simultaneously, while the objective function is to find the optimal backbone $\alpha_b^*$ and the head $\alpha_h^*$ via ranking the accuracy $\emph{Acc}_\emph{val}^\emph{trk}$ of candidate architectures on \emph{validation} set of the tracking data. The evaluation of $\emph{Acc}_\emph{val}^\emph{trk}$ only requires inference because the weights of the architectures $\alpha_b$ and $\alpha_h$ are inherited from $W_b^*(\alpha_b)$ and $W_h^*(\alpha_h)$ (without the need of extra training). 
Note that, before starting the supernet training, we use the pre-trained weight $W_b^p$ to initialize the parameter $W_b$, \emph{i.e.}, $W_b \leftarrow W_b^p$, which speeds up convergence while improving tracking performance. 
During search, it is unaffordable to rank the accuracy of all the architectures in search space, the same as previous work~\cite{SPOS,DetNAS}, we resort to evolutionary algorithms~\cite{real2019regularized,SPOS} to find the most promising one.

\emph{Architecture Constraints}. 
In real-world deployments, object trackers are usually required to satisfy additional constraints, such as memory footprint, model Flops, energy consumption, etc. 
In our method, we mainly consider the model size and Flops, which are two key indicators when evaluating whether a tracker can be deployed on specific resource-constrained devices. 
We preset budgets on networks' \emph{Params} and \emph{Flops} and impose constraints as 
\begin{equation}
	\label{eqn4}
	\begin{aligned}
		Flops(\alpha_b^*) + Flops(\alpha_h^*) ~&\leq~ Flops_{max},\\
		Params(\alpha_b^*) + Params(\alpha_h^*) ~&\leq~ Params_{max}.
	\end{aligned}
\end{equation}
The evolutionary algorithm is flexible in dealing with different budget constraints, because the mutation and crossover processes can be directly controlled
to generate proper candidates to satisfy the constraints~\cite{SPOS}. Search can also be repeated many times on the same supernet once trained, using different constraints 
(e.g., \emph{Flops}$_{max}$ = 600M or others). These properties naturally make one-shot paradigm practical and effective for searching tracking architectures specialized to diverse deployment scenarios.

\begin{table}[t]
	\centering
	\caption{Search space and supernet structure. 
	``$N_{choices}$" represents the number of choices for the current block. ``$Chn$" and ``$Rpt$" denote the number of channels per block and the maximum number of repeated blocks in a group, respectively. ``Stride" indicates the convolutional stride of the first block in each repeated group.
	The classification and regression heads are allowed to use different numbers of channels, denoted as $C_1,C_2\in\{128,192,256\}$. The input is a search image with size of 256$\times$256$\times$3.
	} 
	\vspace{-1mm}
	\resizebox{\columnwidth}{!}{
		\begin{tabular}{c|cccccc}
			\toprule[1.2pt]
			&Input Shape&Operators&$N_{choices}$&$Chn$&$Rpt$&Stride\\
			\midrule[1.2pt]
			\multirow{6}*{\rotatebox{90}{Backbone}}&$256^2\times3$  & $3\times3$ Conv & 1 & 16 & 1 & 2\\
			&$128^2\times16$  & DSConv & 1 & 16 & 1 & 1\\
			&$128^2\times16$ & MBConv & 6 & 24 & 2 & 2\\
			&$64^2\times24$ & MBConv & 6 & 40 & 4 & 2\\
			&$32^2\times40$ & MBConv & 6 & 80 & 4 & 2\\
			&$16^2\times80$ & MBConv & 6 & 96 & 4 & 1\\
			\midrule[1.2pt]
			\multirow{3}{*}{\small \rotatebox{90}{Cls Head}}&$16^2\times128$&DSConv&6&$C_1$&1&1\\
			&$16^2\times C_1$&DSConv / Skip&3&$C_1$&7&1\\
			&$16^2\times C_1$&3x3 Conv&1&1&1&1\\
			\midrule[1.2pt]
			\multirow{3}{*}{\small \rotatebox{90}{Reg Head}}&$16^2\times128$&DSConv&6&$C_2$&1&1\\
			&$16^2\times C_2$&DSConv / Skip&3&$C_2$&7&1\\
			&$16^2\times C_2$&3x3 Conv&1&4&1&1\\
			\bottomrule[1.2pt]
		\end{tabular}
	}
	\label{tab:design}
\vspace{-3mm}
\end{table}	

\subsection{Search Space}
To search for efficient neural architectures, we use depthwise separable convolutions (DSConv)~\cite{DSConv} and mobile inverted bottleneck (MBConv)~\cite{Mobilenetv2} with squeeze-excitation module~\cite{SENet,MobileNetv3} to construct a new search space. 
The space is composed of a backbone part~$\mathcal{A}_b$ and a head part~$\mathcal{A}_h$, which are elaborated in Tab.~\ref{tab:design}. 

\begin{figure*}[!t]
\begin{center}
\includegraphics[width=1\linewidth]{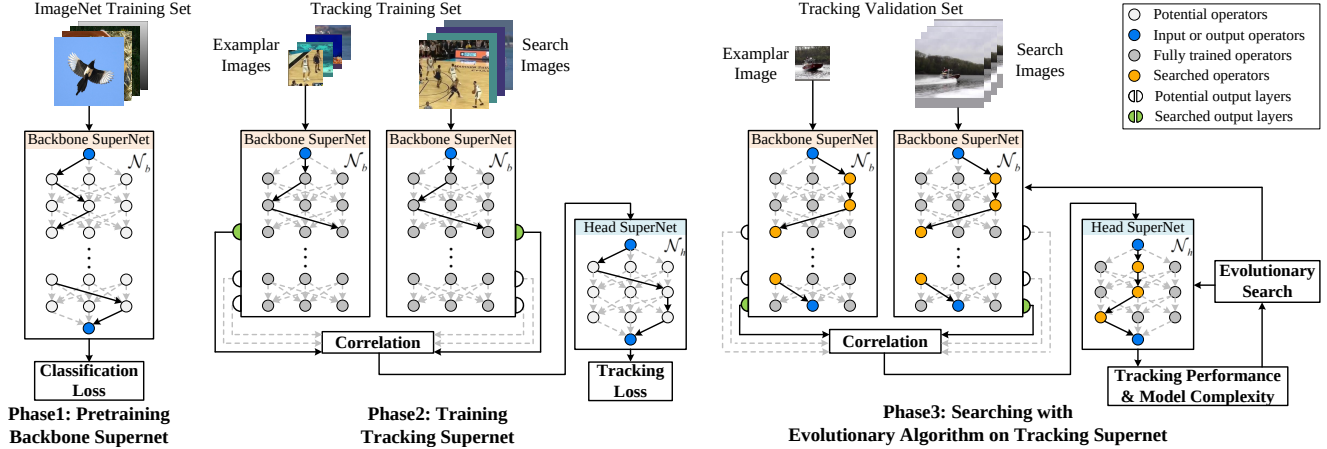}
\end{center}
\vspace{-4mm}
\caption{Search pipeline of the proposed LightTrack. There are three phases: pretraining backbone supernet, training tracking supernet, and searching with evolutionary algorithm on the tracking supernet. Better view in color with zoom-in.}
\label{fig-pipeline}
\vspace{-3mm}
\end{figure*}

{\textbf{Backbone Space $\mathcal{A}_b$}.} There are six basic building blocks in the backbone space, including MBConv with kernel sizes of $\{3,5,7\}$ and expansion rates of $\{4,6\}$. 
Backbone candidates are constructed by stacking the basic blocks. All candidates  in the space have $4$ stages with a total stride of $16$. In each stage, the first block has a stride of 2 for feature downsampling. 
Except for the first two stages, each stage contains up to 4 blocks for search.
There are $14$ layers in the backbone space, as listed in Tab.~\ref{tab:design} (\emph{i.e.}, the layers with a choice number of 6). This space contains about $6^{14}$$\approx$7.8$\times$$10^{10}$ possible backbone architectures for search.

\textbf{Head Space $\mathcal{A}_h$}. A head architecture candidate contains two branches: one for classification while the other for regression. Both of them include at most 8 searchable layers (see Tab.~\ref{tab:design}). 
The first layer is a DSConv with kernel sizes of $\{3,5\}$ and channel numbers of $\{128,192,256\}$. The subsequent $7$ layers follow the same channel setting as the first layer, and have kernel choices of $\{3,5\}$. An additional skip connection is used to enable elastic depth of head architectures~\cite{NASRL}. Different from the backbone space, the head does not include the kernel choice of $7$ because the feature resolution has been relatively low. 
The head space contains about  ${(3\times3^{8})}^2$$\approx$3.9$\times10^{8}$ possible architectures for search. 

In addition, at present, there is no definitive answer to the question of which layer's feature is more suitable for object tracking. We thereby add a new dimension in the search space to allow the one-shot method to determine the output feature layer automatically. Specifically, during supernet training, we randomly pick up an end layer from the last eight blocks in the backbone supernet, and use the output of the picked layer as the extracted feature. Such strategy is able to sample different possible blocks, and allows evolutionary search algorithm to evaluate which layer is better.

It is worth noting that the defined search space contains architectures ranging from 208M to 1.4G Flops with parameter sizes from 0.2M to 5.4M. Such space is much more lightweight than existing handcrafted networks. For example, the human-designed SiamRPN++ with ResNet-50 backbone has 48.9G FLOPs with 54M Params~\cite{ResNet}, being orders of magnitude more complex than architectures in the designed search space. This low-complexity space makes the proposed one-shot NAS algorithm easier to find promising lightweight architectures for tracking.

\newcommand{\tabincell}[2]{\begin{tabular}{@{}#1@{}}#2\end{tabular}}  
\begin{table*}[!th]
\small
    \centering
    \caption{\small
 Comparisons on VOT-19~\cite{VOT2019}. (G) and (M) represent using GoogleNet and MobileNet-V2 as backbones, respectively. DiMP$^r$ indicates the real-time version of DiMP, as reported in ~\cite{VOT2019}. Ocean(off) denotes the offline version of Ocean~\cite{Ocean}. Some values are missing because either the tracker is not open-resourced or the online update module does not support precise Flops estimation.}
    \vspace{-1mm}
    \resizebox{\linewidth}{!}{
    \begin{tabular}{c|ccccccccccc}
        \hline
        &\tabincell{c}{SiamMask\\~\cite{SiamMask}}&\tabincell{c}{SiamFC++(G)\\~\cite{SiamFC++}}&\tabincell{c}{SiamRPN++(M)\\~\cite{SiamRPNplusplus}}&\tabincell{c}{ATOM\\~\cite{ATOM}}&\tabincell{c}{TKU\\~\cite{TKU}}&\tabincell{c}{DiMP$^r$\\~\cite{DiMP}}&\tabincell{c}{Ocean(off)\\~\cite{Ocean}}&\tabincell{c}{\textbf{Ours}\\Mobile}&\tabincell{c}{\textbf{Ours}\\LargeA}&\tabincell{c}{\textbf{Ours}\\LargeB}\\
        \hline
        EAO($\uparrow$)&0.287&0.288&0.292&0.301&0.314&0.321&0.327&\textbf{\textcolor[rgb]{0,0,1}{0.333}}&\textbf{\textcolor[rgb]{0,1,0}{0.340}}&\textbf{\textcolor[rgb]{0.8,0.2,0}{0.357}} \\
        Accuracy($\uparrow$)&\textbf{\textcolor[rgb]{0,1,0}{0.594}}&0.583&0.580&\textbf{\textcolor[rgb]{0.8,0.2,0}{0.603}}&0.589&0.582&\textbf{\textcolor[rgb]{0,0,1}{0.590}}&0.536&0.540&0.552 \\
        Robustness($\downarrow$)&0.461&0.406&0.446&0.411&0.349&0.371&0.376&\textbf{\textcolor[rgb]{0,0,1}{0.321}}&\textbf{\textcolor[rgb]{0,1,0}{0.315}}&\textbf{\textcolor[rgb]{0.8,0.2,0}{0.310}} \\
        FLOPs(G)($\downarrow$)&15.5&17.5&7.0&-&-&-&20.3&\textbf{\textcolor[rgb]{0.8,0.2,0}{0.53}}&\textbf{\textcolor[rgb]{0,1,0}{0.78}}&\textbf{\textcolor[rgb]{0,0,1}{0.79}} \\
        Parameters(M)($\downarrow$)&16.6&13.9&11.2&8.4&-&26.1&25.9&\textbf{\textcolor[rgb]{0.8,0.2,0}{1.97}}&\textbf{\textcolor[rgb]{0,1,0}{2.62}}&\textbf{\textcolor[rgb]{0,0,1}{3.13}}\\
        \hline
    \end{tabular}
    }
    \label{tab-vot}
\vspace{-3mm}
\end{table*}

\subsection{Search Pipeline}
Our LightTrack includes three sequential phases: pre-training backbone supernet, training tracking supernet, and searching with evolutionary algorithm on the trained supernets. The overall pipeline is visualized in Fig.~\ref{fig-pipeline}.

{\textbf{Phase 1: Pre-training Backbone Supernet}.} The backbone supernet $\mathcal{N}_b$ encodes all possible backbone networks in the search space $\mathcal{A}_b$. The structure of $\mathcal{N}_b$ is presented in Tab.~\ref{tab:design}.
As defined in Eq.~(\ref{eqn_spos_class}), the pre-training of the backbone supernet $\mathcal{N}_b$ is to optimize the cross-entropy loss on ImageNet. To decouple the weights of individual subnets, we perform uniform path sampling for the pre-training. In other words, in each batch, only one random path is sampled for feedforward and backward propagation, while other paths are frozen. 

{\textbf{Phase 2: Training Tracking Supernet}.} 
The structure of the tracking supernet $\mathcal{N}$ is visualized in Fig.~\ref{fig-pipeline} (middle). In essence, it is a variant of Siamese tracker~\cite{SiamRPNplusplus,Ocean}.  
Specifically, it takes a pair of tracking images as the input, comprising an exemplar image and a search image. The exemplar image represents the object of interest, while the search image represents the search area in subsequent video frames. Both inputs are processed by the pre-trained backbone network for feature extraction. The generated two feature maps are cross-correlated to generate correlation volumes. The head network contains one classification branch and one regression branch for object localization. The architecture of the head supernet can be found in Tab.~\ref{tab:design}.

The training also adopts the single-path uniform sampling scheme, but involving the tracking head and metrics. In each iteration, the optimizer updates one random path sampled from the backbone and head supernets. The loss function $\mathcal{L}_{train}^{trk}$ in Eq.~(\ref{eqn_spos_track}) includes the common-used binary cross-entropy loss for foreground-background classification and the IoU loss~\cite{iouloss} for object bounding-box regression.

{\textbf{Phase 3: Searching with Evolutionary Algorithm}.} The last phase is to perform evolutionary search on the trained supernet. Paths in the supernet are picked and evaluated under the direction of the evolutionary controller. At first, a population of architectures is initialized randomly. The top-$k$ architectures are picked as parents to generate child networks. The next generation networks are generated by mutation and crossover. For crossover, two randomly selected candidates are crossed to produce a new one. For mutation, a randomly selected candidate mutates its every choice block with probability 0.1 to produce a new candidate. Crossover and mutation are repeated to generate enough new candidates that meet the given architecture constraints in Eq.(\ref{eqn4}). 

One necessary detail is about Batch Normalization~\cite{BatchNormalization}. During search, subnets are sampled in a random way from the supernets. The issue is that the batch statistics on one path should be independent of others~\cite{SPOS,DetNAS}. Therefore, we need to recalculate batch statistics for each single path (subnet) before inference. We sample a random subset from the tracking training set to
recompute the batch statistics for the single path to be evaluated. 
It is extremely fast and takes only a few seconds because no back-propagation is involved.


\begin{table*}[!th]
\small

    \centering

    \caption{
    \small
    Comparisons on GOT-10k~\cite{GOT10K}. (R) and (G) represents ResNet-50 and GoogleNet, respectively.}
    \vspace{-2mm}
    \resizebox{\linewidth}{!}{
    \begin{tabular}{c|cccccccccc}
        \hline
  \small
        &\tabincell{c}{DaSiam\\~\cite{DSiam}}&\tabincell{c}{SiamRPN++(R)\\~\cite{SiamRPNplusplus}}&\tabincell{c}{ATOM\\~\cite{ATOM}}&\tabincell{c}{Ocean-offline\\~\cite{Ocean}}&\tabincell{c}{SiamFC++(G)\\~\cite{SiamFC++}}&\tabincell{c}{Ocean-online\\~\cite{Ocean}} &\tabincell{c}{DiMP-50\\~\cite{DiMP}}&\tabincell{c}{\textbf{Ours}\\Mobile}&\tabincell{c}{\textbf{Ours}\\LargeA}&\tabincell{c}{\textbf{Ours}\\LargeB}\\
        \hline
        AO($\uparrow$)&0.417&0.518&0.556&0.592&0.595&\textbf{\textcolor[rgb]{0,0,1}{0.611}}&\textbf{\textcolor[rgb]{0,0,1}{0.611}}&\textbf{\textcolor[rgb]{0,0,1}{0.611}}&\textbf{\textcolor[rgb]{0,1,0}{0.615}}&\textbf{\textcolor[rgb]{0.8,0.2,0}{0.623}} \\
        SR0.5($\uparrow$)&0.461&0.618&0.634&0.695&0.695&\textbf{\textcolor[rgb]{0,0,1}{0.721}}&0.712&0.710&\textbf{\textcolor[rgb]{0,1,0}{0.723}}&\textbf{\textcolor[rgb]{0.8,0.2,0}{0.726}} \\
        FLOPs(G)($\downarrow$)&21.0&48.9&-&20.3&17.5&-&-&\textbf{\textcolor[rgb]{0.8,0,0}{0.53}}&\textbf{\textcolor[rgb]{0,1,0}{0.78}}&\textbf{\textcolor[rgb]{0,0,1}{0.79}} \\
        Parameters(M)($\downarrow$)&19.6&54.0&8.4&25.9&13.9&44.3&26.1&\textbf{\textcolor[rgb]{0.8,0.2,0}{1.97}}&\textbf{\textcolor[rgb]{0,1,0}{2.62}}&\textbf{\textcolor[rgb]{0,0,1}{3.13}}\\
        \hline
    \end{tabular}
    }
    \label{tab-got10k}
\vspace{-3mm}
\end{table*}

\begin{table*}[!th]
\small
    \centering
    \caption{\small Comparisons on TrackingNet \textit{test} set~\cite{trackingnet}. (A) and (R) represent AlexNet and ResNet-50, respectively.}
    \vspace{-2mm}
    \resizebox{\linewidth}{!}{
    \begin{tabular}{c|ccccccccccc}
        \hline
\small
        &\tabincell{c}{RTMDNet\\~\cite{RTMDNet}}&\tabincell{c}{ECO\\~\cite{ECO}}&\tabincell{c}{DaSiam\\~\cite{DSiam}}&\tabincell{c}{C-RPN\\~\cite{CascadedSiameseRPN}}&\tabincell{c}{ATOM\\~\cite{ATOM}}&\tabincell{c}{SiamFC++(A)\\~\cite{SiamFC++}}&\tabincell{c}{SiamRPN++(R)\\~\cite{SiamRPNplusplus}}&\tabincell{c}{DiMP-50\\~\cite{DiMP}}&\tabincell{c}{\textbf{Ours}\\Mobile}&\tabincell{c}{\textbf{Ours}\\LargeA}&\tabincell{c}{\textbf{Ours}\\LargeB}\\
        \hline
        P(\%)&53.3&55.9&59.1&61.9&64.8&64.6&69.4&68.7&\textbf{\textcolor[rgb]{0,0,1}{69.5}}&\textbf{\textcolor[rgb]{0,1,0}{70.0}}&\textbf{\textcolor[rgb]{0.8,0.2,0}{70.8}}\\
        $P_{norm}$(\%)&69.4&71.0&73.3&74.6&77.1&75.8&\textbf{\textcolor[rgb]{0,1,0}{80.0}}&\textbf{\textcolor[rgb]{0.8,0.2,0}{80.1}}&77.9&78.8&\textbf{\textcolor[rgb]{0,0,1}{78.9}}\\
        AUC(\%)&58.4&61.2&63.8&66.9&70.3&71.2&\textbf{\textcolor[rgb]{0,0,1}{73.3}}&\textbf{\textcolor[rgb]{0.8,0.2,0}{74.0}}&72.5&\textbf{\textcolor[rgb]{0,1,0}{73.6}}&\textbf{\textcolor[rgb]{0,0,1}{73.3}}\\
        \hline
    \end{tabular}
    }
    \label{tab-trackingnet}
\vspace{-3mm}
\end{table*}

\section{Experiments}\label{Sec5}
\subsection{Implementation Details}
\vspace{-1mm}
\textbf{Search.} Following the search pipeline, we first pre-train the backbone supernet on ImageNet for 120 epochs using the following settings: SGD optimizer with momentum 0.9 and weight decay 4e-5, initial learning rate 0.5 with linear annealing. Then, we train the head and the backbone supernets jointly on tracking data. The same as previous work~\cite{Ocean}, the tracking data consists of Youtube-BB~\cite{Youtube}, ImageNet VID~\cite{ImageNet}, ImageNet DET~\cite{ImageNet}, COCO~\cite{COCO} and the training split of GOT-10K~\cite{GOT10K}. 
The training takes 30 epochs, and each epoch uses $6\times10^5$ image pairs. The whole network is optimized using SGD optimizer with momentum 0.9 and weight decay 1e-4. Each GPU hosting 32 images, hence the mini-batch size is 256 images per iteration. The global learning rate increases linearly from 1e-2 to 3e-2 during the first 5 epochs and decreases logarithmically from 3e-2 to 1e-4 in the rest epochs. We freeze the parameters of the backbone in the first 10 epochs and set their learning rate to be $10\times$ smaller than the global learning rate in the rest epochs. Finally, to evaluate the performance of paths in the supernet, we choose the validation set of GOT-10K~\cite{GOT10K} as the evaluation data, since it does not have any overlap with both the training and the final test data.  

\textbf{Retrain.} After evolutionary search, 
we first retrain the discovered backbone network for 500 epochs on Imagenet using similar settings as EfficientNet~\cite{EfficientNet}: MSProp optimizer with momentum 0.9 and decay 0.9, weight decay 1e-5, dropout ratio 0.2, initial learning rate 0.064 with a warmup in the first 3 epochs and a cosine annealing, AutoAugment~\cite{Autoaugment} policy and exponential moving average are adopted for training. Next, we fine-tune the discovered backbone and head networks on the tracking data. The fine-tuning settings in this step are similar to those of the supernet fine-tuning. The main differences include two aspects. 1) The searched architecture is trained for 50 epochs, which is longer than that of the tracking supernet fine-tuning. (2) The global learning rate increases from 2e-2 to 1e-1 during the first 5 epochs and then decreases from 1e-1 to 2e-4 in the rest epochs.

\textbf{Test.} The inference follows the same protocols as in ~\cite{SiameseFC,SiameseRPN}. The feature of the target object is computed once at the first frame, and then consecutively matched with subsequent search images. The hyper-parameters in testing are selected with the tracking toolkit~\cite{Ocean}, which contains an automated parameter tuning algorithm. Our trackers are implemented using Python 3.7 and PyTorch 1.1.0. The experiments are conducted on a server with 8 Tesla V100 GPUs and a Xeon E5-2690 2.60GHz CPU.


\vspace{-1mm}
\subsection{Results and Comparisons}
\vspace{-2mm}
We compare LightTrack to existing hand-designed object trackers with respect to model performance, complexity and run-time speed. The performance is evaluated on four benchmarks, including VOT-19~\cite{VOT2019}, GOT-10K~\cite{GOT10K}, TrackingNet~\cite{trackingnet} and LaSOT~\cite{LaSOT}, while the speed is tested on resource-constrained hardware platforms, 
involving Apple iPhone7 PLUS, Huawei Nova 7 5G, and Xiaomi Mi 8. Moreover, we provide three versions of LightTrack under different resource constraints, \emph{i.e.}, LightTrack Mobile ($\leq$600M Flops, $\leq$2M Params), LargeA ($\leq$800M Flops, $\leq$3M Params) and LargeB ($\leq$800M Flops, $\leq$4M Params).

{\textbf{VOT-19.}} This benchmark contains 60 challenging sequences, and measures tracking accuracy and robustness simultaneously by expected average overlap (EAO). As reported in Tab.~\ref{tab-vot}, 
LightTrack-Mobile achieves superior performance compared to existing SOTA offline trackers, such as SiamRPN++~\cite{SiamRPNplusplus} and SiamFC++~\cite{SiamFC++}, while using $>$10 times fewer model Flops and Params. 
Furthermore, compared to the trackers with online update, such as ATOM~\cite{ATOM} and DiMP$^r$~\cite{DiMP}, LightTrack-LargeB is also competitive, surpassing them by 5.6\% and 3.6\% respectively. This demonstrates the efficacy of the proposed one-shot search algorithm and the discovered architecture.

{\textbf{GOT-10K.}} GOT-10K~\cite{GOT10K} is a new benchmark covering a wide range of common challenges in object tracking, such as deformation and occlusion. Tab.~\ref{tab-got10k} shows that LightTrack obtains state-of-the-art performance, compared to current prevailing trackers. The AO score of LightTrack-Mobile is 1.6\% and 1.9\% superior than SiamFC++(G)~\cite{SiamFC++} and Ocean(off)~\cite{Ocean}, respectively. Besides, if we loosen the computation constraint, the performance of LightTrack will be further improved. For example, LightTrack-LargeB outperforms DiMP-50~\cite{DiMP} by 1.2\%, while using 8$\times$ fewer Params (3.1 \emph{v.s.} 26.1 M).

\begin{figure}[!t]
  \begin{center}
  \begin{tabular}{cc}
  \hspace{-5mm}\includegraphics[width=0.5\linewidth]{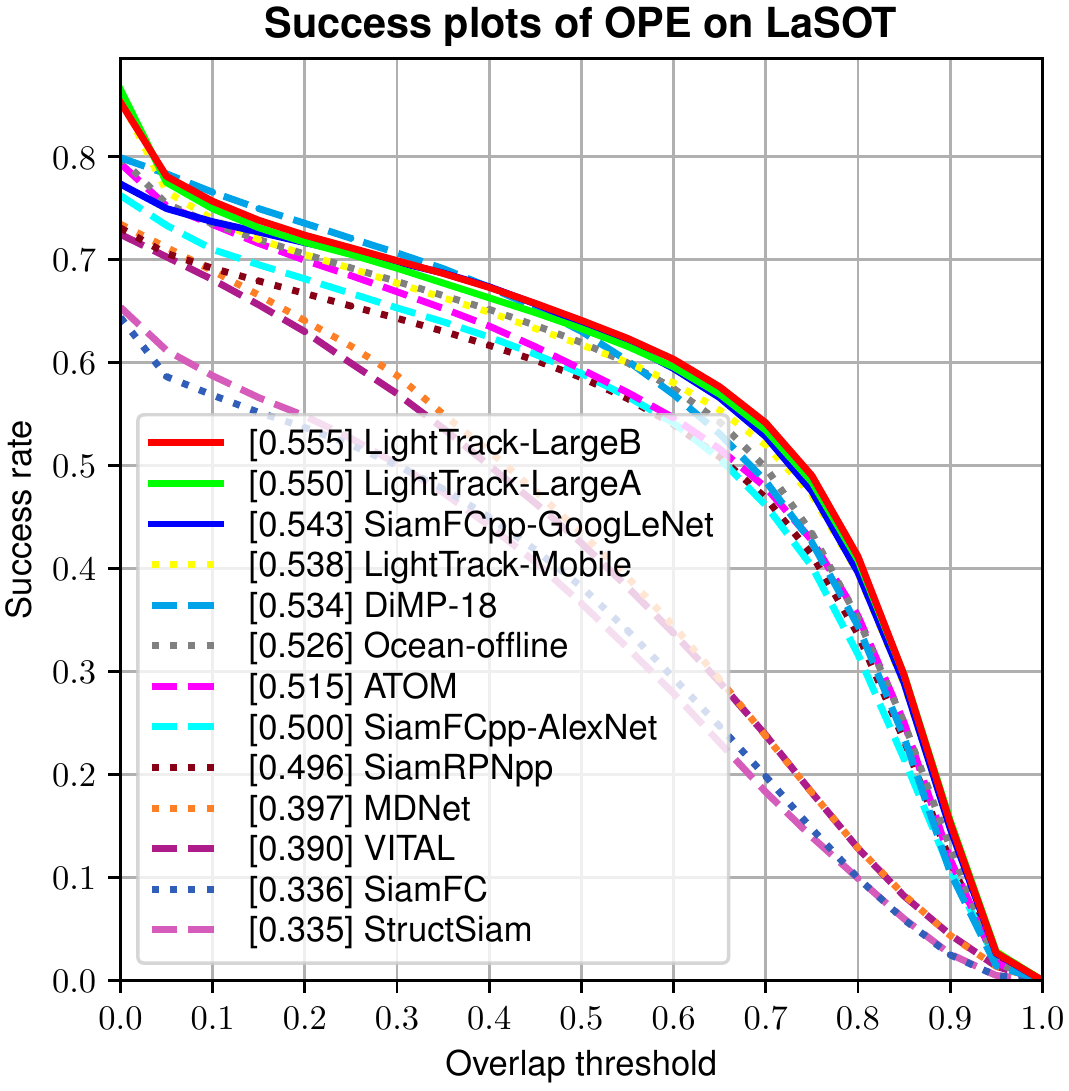} \ &\hspace{-5mm}
  \includegraphics[width=0.5\linewidth]{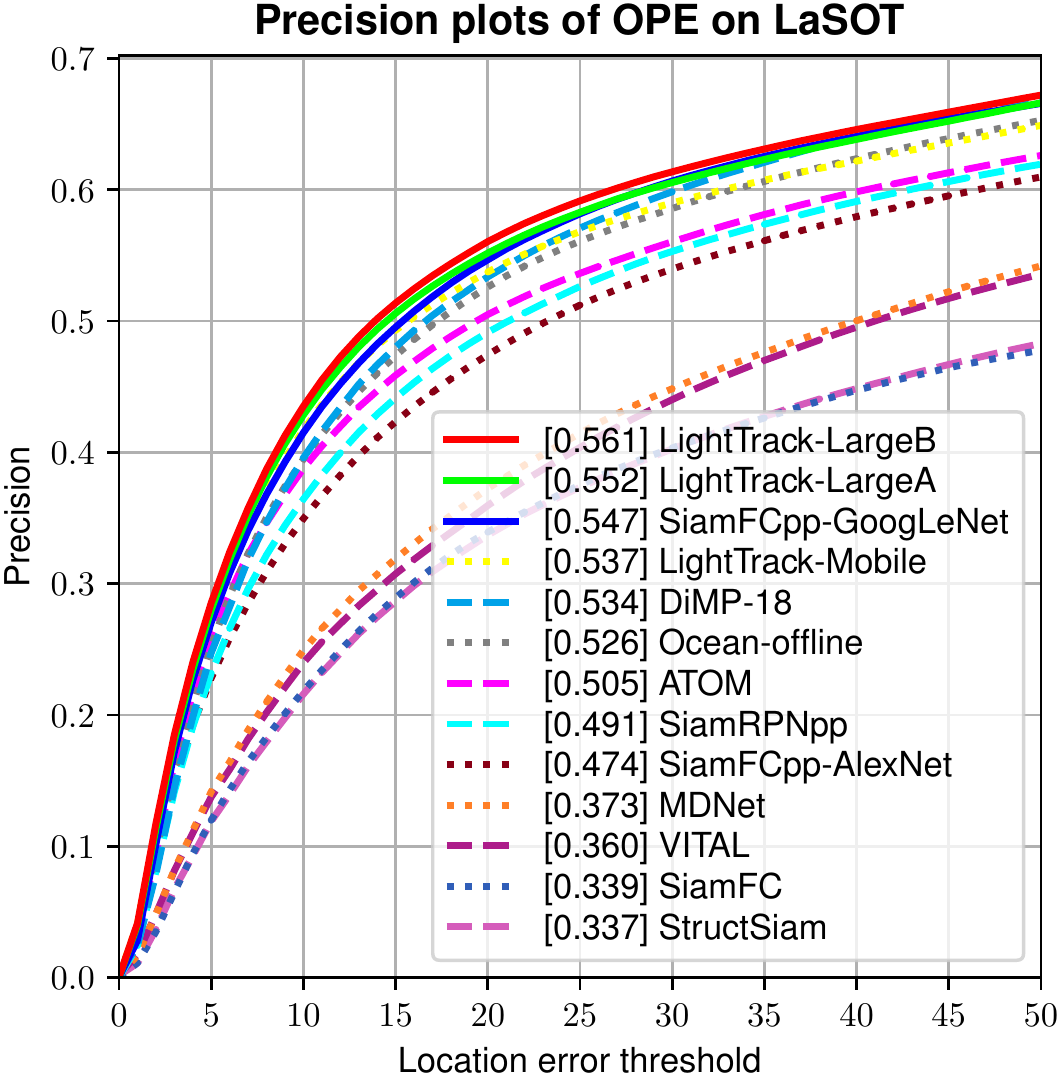}\\
  \end{tabular}
  \end{center}
  \vspace{-5mm}
  \caption{Comparisons on LaSOT \emph{test} dataset~\cite{LaSOT}.}
  \label{fig-lasot}
\vspace{-4.5mm}
\end{figure}

{\textbf{TrackingNet.}} TrackingNet~\cite{trackingnet} is a large-scale short-term tracking benchmark containing 
511 video sequences in \emph{test} set. Tab.~\ref{tab-trackingnet} presents that LightTrack-Mobile achieves better precision (69.5\%), being 0.8\% higher than DiMP-50~\cite{DiMP}. Besides, the $P_{norm}$ and AUC of LightTrack-Mobile are comparable to SiamRPN++ and DiMP-50, while using 96\% and 92\% fewer model Params, respectively.  

\vspace{-1mm}
{\textbf{LaSOT.}} LaSOT~\cite{LaSOT} is by far the largest  single object tracking benchmark with high-quality frame-level annotations. As shown in Fig.~\ref{fig-lasot}, LightTrack-LargeB achieves a success score of 0.555, which surpasses 
SiamFC++(G)~\cite{SiamFC++} 
and Ocean-offline~\cite{Ocean} by 1.2\% and 2.9\%, respectively. Compared to the online DiMP-18~\cite{DiMP}, LightTrack-LargeB improves the success score by 2.1\%, while using 12$\times$ fewer Params (3.1 \emph{v.s.} 39.3 M).

\begin{figure}[!t]
\begin{center}
\includegraphics[height=0.525\linewidth]{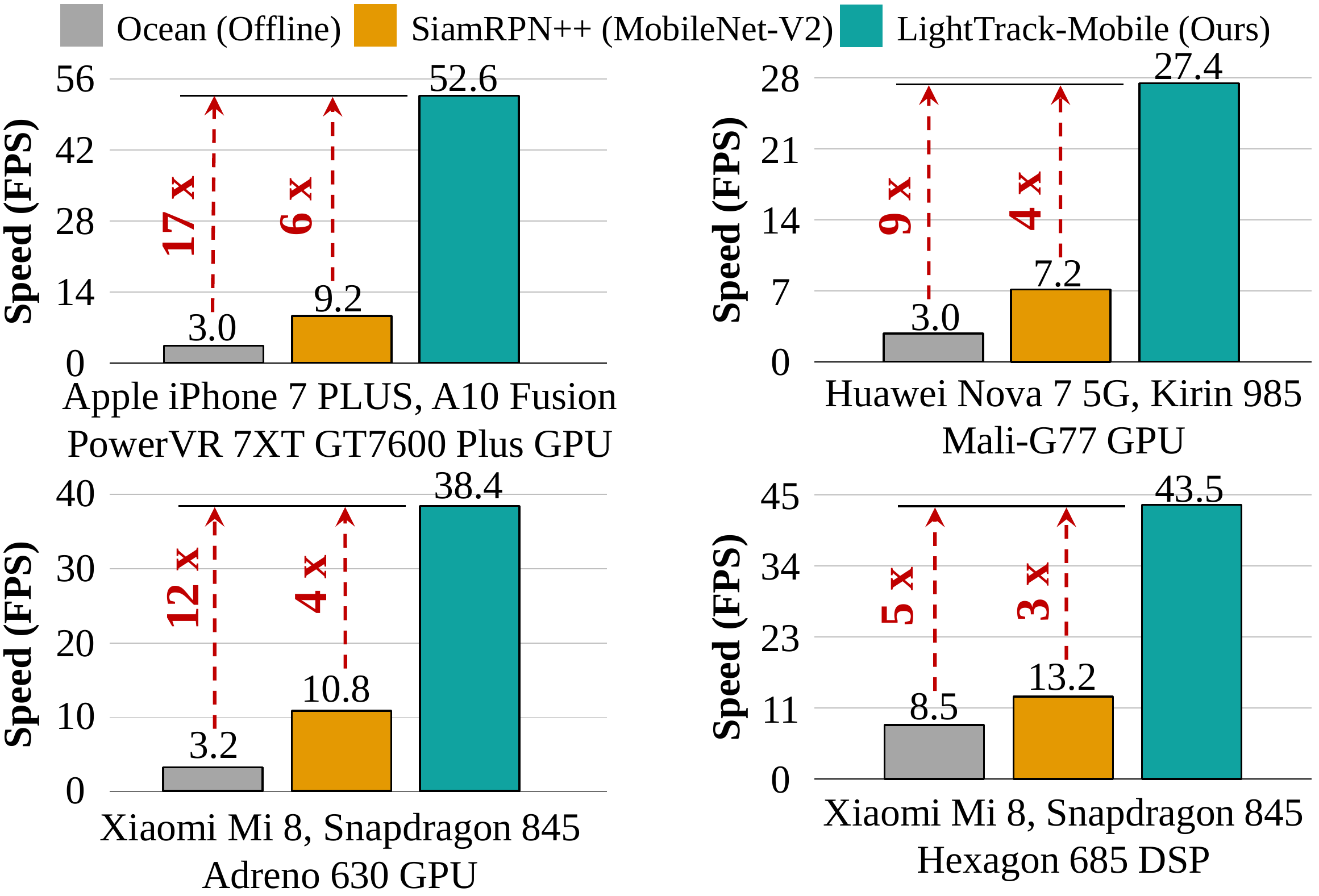}
\end{center}
\vspace{-5mm}
\caption{Run-time speed on resource-limited platforms.}
\label{fig-speed}
\vspace{-4.5mm}
\end{figure}

{\textbf{Speed.}} Fig.~\ref{fig-speed} summarizes the run-time speed of LightTrack on resource-limited mobile  platforms, , including Apple iPhone 7 Plus, Huawei Nova 7 and Xiaomi Mi 8. We observe that SiamRPN++~\cite{SiamRPNplusplus} and Ocean~\cite{Ocean} cannot run at real-time speed (\emph{i.e.}, $<$ 25 \emph{fps}) on these edge devices, such as Snapdragon 845 Adreno 630 GPU and Hexagon 685 DSP. In contrast, our LightTrack run much more efficiently, being {\textbf{3$\sim$6$\times$ faster}} than SiamRPN++ (MobileNetV2 backbone), and {\textbf{5$\sim$17$\times$ faster}} than Ocean (offline) on Snapdragon 845 GPU and DSP~\cite{snapdragon}, Apple A10 Fusion PowerVR GPU~\cite{Apple_A10}, and Kirin 985 Mali-G77 GPU~\cite{kirin985}. The  real-time speed allows LightTrack to be deployed and applied in resource-constrained applications, such as camera drones where edge chipsets are commonly used. The speed improvements also demonstrate that LightTrack is effective and can find more compact and efficient object trackers.

\begin{figure*}[!t]
\begin{center}
\vspace{-4mm}
\includegraphics[width=1\linewidth]{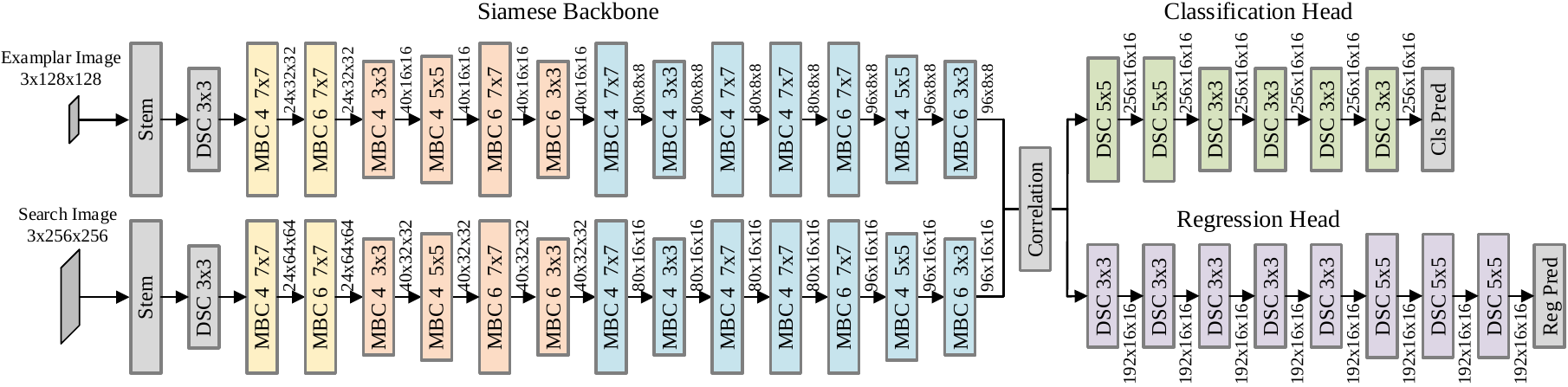}
\end{center}
\vspace{-4.5mm}
\caption{The architecture searched by the proposed LightTrack (Mobile). The searchable layers are drawn in colors while the fixed/pre-defined parts are plotted in grey. The ``Stem'' consists of a normal 2D convolution layer with kernel size of 3$\times$3 and stride of 2, a BatchNorm layer, and a Swish activation layer. ``DSConv'' indicates depthwise separable convolution \cite{DSConv} while ``MBConv'' denotes mobile inverted bottleneck~\cite{Mobilenetv2} with squeeze excitation~\cite{SENet}.} 
\label{fig-arch}
\vspace{-3mm}
\end{figure*}


\subsection{Ablation and Analysis}
\emph{Component-wise Analysis.} We evaluate the effects of different components in our LightTrack on VOT-19~\cite{VOT2019}, and report the results
in Tab.~\ref{tab-ablation}. Our baseline is a handcrafted mobile tracker, which takes MobileNetV3-large~\cite{MobileNetv3} as the backbone (chopping off the last stage), and outputs features from the last layer with a stride of 16. The head network stacks 8 layers of depthwise separable convolution (DSConv)~\cite{DSConv} in both classification and regression branches. For each DSConv, the kernel size is set to 3$\times$3 and the number of channels is 256. The EAO performance of the baseline is 0.268. 
For ablation, we add the components in the baseline into search and change the handcrafted architectures with automatically searched ones. 
%
As presented in Tab.~\ref{tab-ablation} \text{\#}2, when the backbone architecture is automatically searched, the EAO performance is improved by 2.4\%. This demonstrates that the hand-designed MobileNetV3-large backbone is not optimal for object tracking, because it is primarily designed for image classification, where the precise localization of the object is not paramount. If we add the output feature layer into search, the performance is further improved to 0.307. This shows that our method can search out a better layer for feature extraction. The comparison between \text{\#4} and \text{\#1}  shows that the searchable head architecture is superior to the handcrafted one, inducing 2.9\% EAO gains. When searching the three components together, as shown in \text{\#5}, the complete LightTrack achieves better performance than only searching parts of the tracking network. 

{\textit{Impact of ImageNet Pre-training.}} We pre-train the searched architecture on ImageNet for 0, 200 and 500 epochs, and evaluate their impact for final tracking performance.  As reported in Tab.~\ref{tab-imagenet}, no pre-training has a significantly negative impact on tracking accuracy. Better pre-training allows the tracker to achieve higher performance. 

{\textit{Analysis of Searched Architecture.}} 
Fig.~\ref{fig-arch} visualizes the LightTrack-Mobile architecture searched by the proposed one-shot NAS method. 
We observe several interesting phenomena. 
1) There are about 50\% of the backbone blocks using MBConv with kernel size of 7x7. The underlying reason may be that large receptive fields can improve the localization precision. 
2) The searched architecture chooses the second-last block as the feature output layer.
This may reveals that tracking networks might not prefer high-level features.
3) The classification branch contains fewer layers than the regression branch. 
This may be attributed to the fact that coarse object localization is relatively easier than precise bounding box regression. These findings might enlighten future works on designing new tracking networks. 

\begin{table}[!t]
\small
    \centering
    \caption{\small Ablation for searchable components. \cmark\ indicates automatically searched, while \xmark\ denotes hand-designed. }
    \vspace{-1mm}
    \begin{tabular}{c|ccc|c}
        \hline
        \text{\#}&Backbone&Output Layer&Head&EAO\\
        \hline
        1&\xmark&\xmark&\xmark&0.268\\
        2&\cmark&\xmark&\xmark&0.292\\
        3&\cmark&\cmark&\xmark&0.307\\
        4&\xmark&\xmark&\cmark&0.297\\
        5&\cmark&\cmark&\cmark&0.333\\
        \hline
    \end{tabular}
    \label{tab-ablation}
\vspace{-1mm}
\end{table}

\begin{table}[!t]
\small
    \centering
    \caption{\small Impact of ImageNet Pre-training.}
    \vspace{-2mm}
    \small
    \begin{tabular}{c|ccc}
        \hline
        &Epoch 0&Epoch 200&Epoch 500\\
        \hline
        Top-1 Acc (\%)& -- &72.4&77.6\\
        EAO on VOT-19 (\%)&21.3&31.2&33.3\\
        \hline
    \end{tabular}
    \label{tab-imagenet}
\vspace{-3mm}
\end{table}

\vspace{-1mm}
\section{Conclusion}\label{Sec6}
This paper makes the first effort on designing lightweight object trackers via neural architecture search. The proposed method, \emph{i.e.}, LightTrack, reformulates one-shot NAS specialized for object tracking, as well as introducing an effective search space. Extensive experiments on multiple  benchmarks show that LightTrack achieves state-of-the-art performance, while using much fewer Flops and parameters. Besides, LightTrack can run in real-time on diverse resource-restricted platforms. We expect this work might be able to narrow the gap between academic methods and industrial applications in object tracking field. 

\noindent{\textbf{Acknowledgement. }
We would like to thank the reviewers for their insightful comments. Lu and Wang are supported in part by the National Key R\&D Program of China under Grant No. 2018AAA0102001 and National Natural Science Foundation of China under grant No. 61725202, U1903215, 61829102, 91538201, 61771088, 61751212 and Dalian Innovation leader’s support Plan under Grant No. 2018RD07.}
\newpage

{\small
\bibliographystyle{ieee_fullname}
\bibliography{egbib}

\begin{thebibliography}{10}\itemsep=-1pt

\bibitem{Apple_A10}
\href{https://en.wikipedia.org/wiki/Apple\_A10}{https://en.wikipedia.org/wiki/Apple\_A10}.

\bibitem{kirin985}
\href{https://www.hisilicon.com/en/products/Kirin/Kirin\%20985}{https://www.hisilicon.com/en/products/Kirin/Kirin\%20985}.

\bibitem{snapdragon}
\href{https://www.qualcomm.com/products/snapdragon-845-mobile-platform}{https://www.qualcomm.com/products/snapdragon-845-mobile-platform}.

\bibitem{understand_os}
Gabriel Bender, Pieter-Jan Kindermans, Barret Zoph, Vijay Vasudevan, and Quoc
  Le.
\newblock Understanding and simplifying one-shot architecture search.
\newblock In {\em ICML}, 2018.

\bibitem{SiameseFC}
Luca Bertinetto, Jack Valmadre, Jo{\~a}o~F Henriques, Andrea Vedaldi, and
  Philip H~S Torr.
\newblock Fully-convolutional siamese networks for object tracking.
\newblock In {\em ECCVW}, 2016.

\bibitem{DiMP}
Goutam Bhat, Martin Danelljan, Luc~Van Gool, and Radu Timofte.
\newblock Learning discriminative model prediction for tracking.
\newblock In {\em ICCV}, 2019.

\bibitem{OFA}
Han Cai, Chuang Gan, Tianzhe Wang, Zhekai Zhang, and Song Han.
\newblock Once-for-all: Train one network and specialize it for efficient
  deployment.
\newblock In {\em ICLR}, 2019.

\bibitem{PDARTS}
Xin Chen, Lingxi Xie, Jun Wu, and Qi Tian.
\newblock Progressive differentiable architecture search: Bridging the depth
  gap between search and evaluation.
\newblock In {\em ICCV}, 2019.

\bibitem{DetNAS}
Yukang Chen, Tong Yang, Xiangyu Zhang, Gaofeng Meng, Xinyu Xiao, and Jian Sun.
\newblock {DetNAS}: Backbone search for object detection.
\newblock In {\em NIPS}, 2019.

\bibitem{SiamBAN}
Zedu Chen, Bineng Zhong, Guorong Li, Shengping Zhang, and Rongrong Ji.
\newblock Siamese box adaptive network for visual tracking.
\newblock In {\em CVPR}, 2020.

\bibitem{DSConv}
Fran{\c{c}}ois Chollet.
\newblock Xception: Deep learning with depthwise separable convolutions.
\newblock In {\em CVPR}, 2017.

\bibitem{Autoaugment}
Ekin~D Cubuk, Barret Zoph, Dandelion Mane, Vijay Vasudevan, and Quoc~V Le.
\newblock Autoaugment: Learning augmentation strategies from data.
\newblock In {\em CVPR}, 2019.

\bibitem{ECO}
Martin Danelljan, Goutam Bhat, Fahad~Shahbaz Khan, and Michael Felsberg.
\newblock {ECO}: Efficient convolution operators for tracking.
\newblock In {\em CVPR}, 2017.

\bibitem{ATOM}
Martin Danelljan, Goutam Bhat, Fahad~Shahbaz Khan, and Michael Felsberg.
\newblock {ATOM: Accurate} tracking by overlap maximization.
\newblock In {\em CVPR}, 2019.

\bibitem{Survey}
Thomas Elsken, Jan~Hendrik Metzen, and Frank Hutter.
\newblock Neural architecture search: A survey.
\newblock {\em JMLR}, 20(55):1--21, 2019.

\bibitem{LaSOT}
Heng Fan, Liting Lin, Fan Yang, Peng Chu, Ge Deng, Sijia Yu, Hexin Bai, Yong
  Xu, Chunyuan Liao, and Haibin Ling.
\newblock {LaSOT}: A high-quality benchmark for large-scale single object
  tracking.
\newblock In {\em CVPR}, 2019.

\bibitem{CascadedSiameseRPN}
Heng Fan and Haibin Ling.
\newblock Siamese cascaded region proposal networks for real-time visual
  tracking.
\newblock In {\em CVPR}, 2019.

\bibitem{GCT}
Junyu Gao, Tianzhu Zhang, and Changsheng Xu.
\newblock Graph convolutional tracking.
\newblock In {\em CVPR}, 2019.

\bibitem{NAS-FPN}
Golnaz Ghiasi, Tsung-Yi Lin, and Quoc~V Le.
\newblock {NAS-FPN}: Learning scalable feature pyramid architecture for object
  detection.
\newblock In {\em CVPR}, 2019.

\bibitem{SPOS}
Zichao Guo, Xiangyu Zhang, Haoyuan Mu, Wen Heng, Zechun Liu, Yichen Wei, and
  Jian Sun.
\newblock Single path one-shot neural architecture search with uniform
  sampling.
\newblock In {\em ECCV}, 2020.

\bibitem{deepcompress}
Song Han, Huizi Mao, and William~J Dally.
\newblock Deep compression: Compressing deep neural networks with pruning,
  trained quantization and huffman coding.
\newblock In {\em ICLR}, 2016.

\bibitem{ResNet}
Kaiming He, Xiangyu Zhang, Shaoqing Ren, and Jian Sun.
\newblock Deep residual learning for image recognition.
\newblock In {\em CVPR}, 2016.

\bibitem{MobileNetv3}
Andrew Howard, Mark Sandler, Grace Chu, Liang-Chieh Chen, Bo Chen, Mingxing
  Tan, Weijun Wang, Yukun Zhu, Ruoming Pang, Vijay Vasudevan, et~al.
\newblock Searching for mobilenetv3.
\newblock In {\em ICCV}, 2019.

\bibitem{SENet}
Jie Hu, Li Shen, and Gang Sun.
\newblock Squeeze-and-excitation networks.
\newblock In {\em CVPR}, 2018.

\bibitem{GOT10K}
Lianghua Huang, Xin Zhao, and Kaiqi Huang.
\newblock {GOT-10k}: A large high-diversity benchmark for generic object
  tracking in the wild.
\newblock {\em TPAMI}, 2019.

\bibitem{BatchNormalization}
Sergey Ioffe and Christian Szegedy.
\newblock Batch {N}ormalization: Accelerating deep network training by reducing
  internal covariate shift.
\newblock In {\em ICML}, 2015.

\bibitem{RTMDNet}
Ilchae Jung, Jeany Son, Mooyeol Baek, and Bohyung Han.
\newblock {Real-Time} {MDNet}.
\newblock In {\em ECCV}, 2018.

\bibitem{VOT2019}
Matej Kristan, Jiri Matas, Ales Leonardis, et~al.
\newblock The seventh visual object tracking {VOT2019} challenge results.
\newblock In {\em ICCVW}, 2019.

\bibitem{AlexNet}
Alex Krizhevsky, Ilya Sutskever, and Geoffrey~E Hinton.
\newblock Imagenet classification with deep convolutional neural networks.
\newblock In {\em NIPS}, 2012.

\bibitem{SiamRPNplusplus}
Bo Li, Wei Wu, Qiang Wang, Fangyi Zhang, Junliang Xing, and Junjie Yan.
\newblock {SiamRPN++}: {Evolution} of siamese visual tracking with very deep
  networks.
\newblock In {\em CVPR}, 2019.

\bibitem{SiameseRPN}
Bo Li, Junjie Yan, Wei Wu, Zheng Zhu, and Xiaolin Hu.
\newblock High performance visual tracking with siamese region proposal
  network.
\newblock In {\em CVPR}, 2018.

\bibitem{Zuo}
Feng Li, Cheng Tian, Wangmeng Zuo, Lei Zhang, and Ming-Hsuan Yang.
\newblock Learning spatial-temporal regularized correlation filters for visual
  tracking.
\newblock In {\em CVPR}, 2018.

\bibitem{randomNAS}
Liam Li and Ameet Talwalkar.
\newblock Random search and reproducibility for neural architecture search.
\newblock In {\em UAI}, 2019.

\bibitem{TADT}
Xin Li, Chao Ma, Baoyuan Wu, Zhenyu He, and Ming-Hsuan Yang.
\newblock Target-aware deep tracking.
\newblock In {\em CVPR}, 2019.

\bibitem{COCO}
Tsung-Yi Lin, Michael Maire, Serge~J. Belongie, Lubomir~D. Bourdev, Ross~B.
  Girshick, James Hays, Pietro Perona, Deva Ramanan, Piotr Doll{\'a}r, and
  C.~Lawrence Zitnick.
\newblock {Microsoft COCO}: Common objects in context.
\newblock In {\em ECCV}, 2014.

\bibitem{Auto-Deeplab}
Chenxi Liu, Liang-Chieh Chen, Florian Schroff, Hartwig Adam, Wei Hua, Alan~L
  Yuille, and Li Fei-Fei.
\newblock Auto-{D}eeplab: Hierarchical neural architecture search for semantic
  image segmentation.
\newblock In {\em CVPR}, 2019.

\bibitem{DARTS}
Hanxiao Liu, Karen Simonyan, and Yiming Yang.
\newblock {DARTS:} differentiable architecture searc.
\newblock In {\em ICLR}, 2019.

\bibitem{TStracker}
Yuanpei Liu, Xingping Dong, Wenguan Wang, and Jianbing Shen.
\newblock Teacher-students knowledge distillation for siamese trackers.
\newblock {\em arXiv preprint arXiv:1907.10586}, 2019.

\bibitem{trackingnet}
Matthias Muller, Adel Bibi, Silvio Giancola, Salman Alsubaihi, and Bernard
  Ghanem.
\newblock Trackingnet: A large-scale dataset and benchmark for object tracking
  in the wild.
\newblock In {\em ECCV}, 2018.

\bibitem{MDNet}
Hyeonseob Nam and Bohyung Han.
\newblock Learning multi--domain convolutional neural networks for visual
  tracking.
\newblock In {\em CVPR}, 2016.

\bibitem{ENAS}
Hieu Pham, Melody Guan, Barret Zoph, Quoc Le, and Jeff Dean.
\newblock Efficient neural architecture search via parameters sharing.
\newblock In {\em ICML}, 2018.

\bibitem{real2019regularized}
Esteban Real, Alok Aggarwal, Yanping Huang, and Quoc~V Le.
\newblock Regularized evolution for image classifier architecture search.
\newblock In {\em AAAI}, 2019.

\bibitem{Youtube}
Esteban Real, Jonathon Shlens, Stefano Mazzocchi, Xin Pan, and Vincent
  Vanhoucke.
\newblock Youtube-boundingboxes: A large high-precision human-annotated data
  set for object detection in video.
\newblock In {\em CVPR}, 2017.

\bibitem{ImageNet}
Olga Russakovsky, Jia Deng, Hao Su, Jonathan Krause, Sanjeev Satheesh, Sean Ma,
  Zhiheng Huang, Andrej Karpathy, Aditya Khosla, and Michael Bernstein.
\newblock {ImageNet} {Large} scale visual recognition challenge.
\newblock {\em IJCV}, 2015.

\bibitem{Mobilenetv2}
Mark Sandler, Andrew Howard, Menglong Zhu, Andrey Zhmoginov, and Liang-Chieh
  Chen.
\newblock Mobilenetv2: Inverted residuals and linear bottlenecks.
\newblock In {\em CVPR}, 2018.

\bibitem{EfficientNet}
Mingxing Tan and Quoc Le.
\newblock Efficientnet: Rethinking model scaling for convolutional neural
  networks.
\newblock In {\em ICML}, 2019.

\bibitem{SINT}
Ran Tao, Efstratios Gavves, and Arnold W.~M. Smeulders.
\newblock Siamese instance search for tracking.
\newblock In {\em CVPR}, 2016.

\bibitem{TKU}
Ardhendu~Shekhar Tripathi, Martin Danelljan, Luc Van~Gool, and Radu Timofte.
\newblock Tracking the known and the unknown by leveraging semantic
  information.
\newblock In {\em BMVC}, 2019.

\bibitem{SiamRCNN}
Paul Voigtlaender, Jonathon Luiten, Philip~HS Torr, and Bastian Leibe.
\newblock Siam {R-CNN}: Visual tracking by re-detection.
\newblock In {\em CVPR}, 2020.

\bibitem{SiamMask}
Qiang Wang, Li Zhang, Luca Bertinetto, Weiming Hu, and Philip H.~S. Torr.
\newblock Fast online object tracking and segmentation: {A} unifying approach.
\newblock In {\em CVPR}, 2019.

\bibitem{geneticCNN}
Lingxi Xie and Alan Yuille.
\newblock Genetic cnn.
\newblock In {\em ICCV}, 2017.

\bibitem{SiamFC++}
Yinda Xu, Zeyu Wang, Zuoxin Li, Ye Yuan, and Gang Yu.
\newblock {SiamFC++:} towards robust and accurate visual tracking with target
  estimation guidelines.
\newblock In {\em AAAI}, 2020.

\bibitem{ROAM}
Tianyu Yang, Pengfei Xu, Runbo Hu, Hua Chai, and Antoni~B Chan.
\newblock {ROAM}: Recurrently optimizing tracking model.
\newblock In {\em CVPR}, 2020.

\bibitem{iouloss}
Jiahui Yu, Yuning Jiang, Zhangyang Wang, Zhimin Cao, and Thomas Huang.
\newblock Unitbox: An advanced object detection network.
\newblock In {\em ACM MM}, 2016.

\bibitem{Deeper-wider-SiamRPN}
Zhipeng Zhang and Houwen Peng.
\newblock Deeper and wider siamese networks for real-time visual tracking.
\newblock In {\em CVPR}, 2019.

\bibitem{Ocean}
Zhipeng Zhang, Houwen Peng, Jianlong Fu, Bing Li, and Weiming Hu.
\newblock Ocean: Object-aware anchor-free tracking.
\newblock In {\em ECCV}, 2020.

\bibitem{DSiam}
Zheng Zhu, Qiang Wang, Bo Li, Wei Wu, Junjie Yan, and Weiming Hu.
\newblock Distractor-aware siamese networks for visual object tracking.
\newblock In {\em ECCV}, 2018.

\bibitem{NASRL}
Barret Zoph and Quoc~V Le.
\newblock Neural architecture search with reinforcement learning.
\newblock {\em ICLR}, 2017.

\end{thebibliography}
}

\end{document}


\title{EfficientTracker: Searching Efficient Neural Architectures for Visual Tracking}

\author{First Author\\
Institution1\\
Institution1 address\\
{\tt\small firstauthor@i1.org}
\and
Second Author\\
Institution2\\
First line of institution2 address\\
{\tt\small secondauthor@i2.org}
}

\maketitle

\begin{table}[h]
    \vspace{-0.5cm}
	\centering
	\resizebox{\columnwidth}{!}{
	\begin{tabular}{c|cccccc}
		\toprule[1.2pt]
		&\multirow{2}{*}{Input Shape} & \multirow{2}{*}{Operators} & \multirow{2}{*}{Choices} & \multirow{2}{*}{C} & \multirow{2}{*}{Rep}  & \multirow{2}{*}{S} \\
		\\
		\midrule[1.2pt]
		\multirow{6}*{\rotatebox{90}{Backbone}}&$256^2\times3$  & $3\times3$ Conv   & 1 & 16    & 1 & 2\\
		&$128^2\times16$  & DS Conv & 1 & 16    & 1 & 1\\
		&$128^2\times16$ & MBConv & 6 & 24 & 2 & 2\\
		&$64^2\times24$ & MBConv & 6 & 40 & 4 & 2\\
		&$32^2\times40$ & MBConv & 6 & 80 & 4 & 2\\
		&$16^2\times80$ & MBConv & 6 & 96 & 4 & 1\\
		\midrule[1.2pt]
		\multirow{3}{*}{\small \rotatebox{90}{Cls Head}}&$16^2\times128$&DS Conv&6&$C_1$&1&1\\
		&$16^2\times C_1$&DS Conv / Skip&3&$C_1$&7&1\\
		&$16^2\times C_1$&3x3 Conv&1&1&1&1\\
		\midrule[1.2pt]
		\multirow{3}{*}{\small \rotatebox{90}{Reg Head}}&$16^2\times128$&DS Conv&6&$C_2$&1&1\\
		&$16^2\times C_2$&DS Conv / Skip&3&$C_2$&7&1\\
		&$16^2\times C_2$&3x3 Conv&1&4&1&1\\
		\bottomrule[1.2pt]
	\end{tabular}
	}
	\caption{The structure of the supernetwork. The "MBConv" contains $6$ inverted bottleneck residual block MBConv \cite{Mobilenetv2} ( kernel sizes of \{3,5,7\}) with the squeeze and excitation module (expansion rates \{4,6\}). The "Choices" and "C" represents the number of choices and channels respectively in the current operators. The "Rep" represents the maximum number of repeated blocks in a group. The "S" indicates the convolutional stride of the first block in each repeated group. The classification and regression heads are allowed to use different channel number, denoted as $C_1, C_2$. ($C1,C2\in\{128,192,256\}$)} 
	
	\label{tab:design}
\end{table}	

{\small
\bibliographystyle{ieee_fullname}
\bibliography{egbib}
}